\pdfoutput=1

\documentclass[11pt]{article}

\usepackage[preprint]{acl}

\usepackage{times}
\usepackage{latexsym}

\usepackage[T1]{fontenc}

\usepackage[utf8]{inputenc}

\usepackage{microtype}

\usepackage{inconsolata}
\usepackage{caption}
\usepackage{subcaption}
\usepackage{graphicx}
\usepackage{amsmath}

\title{Transformer-based Causal Language Models Perform Clustering}

\author{Xinbo Wu \\
  University of Illinois Urbana-Champaign \quad \\
  \texttt{xinbowu2@illinois.edu} \\\And
  Lav R.\ Varshney \\
  \quad University of Illinois Urbana-Champaign \\
  \texttt{varshney@illinois.edu} \\}

\begin{document}
\maketitle
\begin{abstract}
Even though large language models (LLMs) have demonstrated remarkable capability in solving various natural language tasks, the capability of an LLM to follow human instructions is still a concern. Recent works \citep{RLHF, DPO, IT_survey} have shown great improvements in the instruction-following capability via additional training for instruction-following tasks. However, the mechanisms responsible for effective instruction-following capabilities remain inadequately understood. Here, we introduce a simplified instruction-following task and use synthetic datasets to analyze a Transformer-based causal language model. Our findings suggest that the model learns task-specific information by clustering data within its hidden space, with this clustering process evolving dynamically during learning. We also demonstrate how this phenomenon assists the model in handling unseen instances, and validate our results in a more realistic setting. Furthermore, we present inspired applications regarding pre-training and alignment.      
\end{abstract}

\section{Introduction} \label{sec:introduction}

In recent years, the development of large language models (LLMs) has marked a noteworthy milestone in the field of artificial intelligence and natural language processing due to their remarkable capabilities ~\citep{GPT3,GPT-4,llama}. However, a significant challenge with LLMs is the misalignment between their training objectives and users' intentions. LLMs are trained to optimize next-word prediction on large-scale language data whereas users expect the model to follow their instructions in a helpful and safe manner \citep{IT_survey}. Techniques such as reinforcement learning from human feedback (RLHF) \citep{RLHF}, direct preference optimization (DPO) \citep{DPO}, and instruction tuning \citep{IT_survey} have been proposed to further train LLMs for instruction following, yielding seemingly great instruction-following capabilities. Instruction-following also harkens back to controllable language models \citep{CTRL}. 

Yet, the mechanisms underlying these successful instruction-following capabilities are not well-understood and require specific analysis. Since LLMs have grown exceedingly complex in terms of their parameters and the data they have been trained on, their analysis is extremely challenging. For example, it is difficult to determine which token to focus on for analyzing a potentially lengthy and intricate textual sequence, so as to extract meaningful interpretation. To gain insights into the hidden mechanisms of LLMs, one approach is through carefully designed experiments. Conducting such experiments requires meticulous control over experimental settings and this is challenging due to the complexity of real-world language data, over which experimenters have limited control. 

This limitation inspires us to devise a simplified instruction-following task with a synthetic dataset that we fully control but reflects some key properties of natural language data. This approach mirrors practices in fields such as experimental psychology, where researchers aim to study the complexities of the human mind under simplified task conditions with controlled stimuli. Since the Transformer architecture \citep{Transformer} is commonly used to build LLMs, we aim to perform analysis on a Transformer-based causal language model (CLM) trained for a simplified instruction-following task to see if the model has any inductive bias. 

More specifically, the ability to correctly recognize a learned task may be needed to successfully execute it. We aim to investigate how task-specific information is encoded into the representation space of the Transformer-based CLM trained for instruction-following. One intuitive hypothesis is that the hidden states corresponding to the same task are arranged close together to form a task-specific cluster, reminiscent of functional modules and topographic maps that neuroscientists have discovered in the brain \citep{Maps2, Maps}. The alternative is that the hidden states are scattered without forming clusters, and the Transformer learns mechanisms to identify tasks via these scattered hidden states. In Section~\ref{sec:experiment}, we find experimental evidence supporting the former. 

This leads us to further questions. First, do these task-specific clusters emerge at a certain point during training or gradually evolve? Second, how might task-specific clustering enhance task performance? To dive into these, we must investigate model learning dynamics, but in-depth analysis of LLM training is expensive due to the long training schedule and high computational costs. Our simplified setting reveals its advantage by allowing us to constrain the scope of the tasks such that a relatively small model is able to fully learn the task in a short training schedule. Specifically, we perform clustering analysis on the hidden states of the Transformer model by training it for the instruction-following task and extend this analysis to the entire learning process. We train the model from scratch to isolate it from complicated and potentially noisy pre-training, leaving studies of pre-training impacts for future work. 

In summary, we present a simplified instruction-following task and generate a synthetic dataset to examine a Transformer-based CLM model. Through this simplified framework, we offer evidence suggesting that the model learns task-specific information by organizing data into clusters within its hidden space. Moreover, we show that the clusters evolve dynamically as the model learns. Importantly, we illustrate how this clustering phenomenon aids the model in handling previously unseen instances. Furthermore, we validate our findings from the simplified task in a more realistic setting and showcase two possible applications on pre-training and alignment respectively inspired by the findings of this work.  

\section{Instruction-Following}
\subsection{Preliminaries} \label{sec:preliminary}
We assume a task is a function $f: \mathcal{X} \rightarrow \mathcal{Y}$ and each pair of its input and output $(x, y)$ is a mapping. We define an instruction-following task as anticipating an output $y$ by giving an instruction $I$ and an input ${x}$ such as ``given a location, state its continent. New York City'', where New York City is the input and the output should be North America. Essentially an instruction serves as a prompt that helps to identify a specific task function $f$. We assume an instruction $I$ is sampled from an instruction distribution $\mathcal{I}$ and instructions sampled from different distributions may be associated with the same task. However, the opposite does not hold since this will lead to an ill-defined problem. 

The input can be either integrated into the instruction or separated out. For simplicity, we use a separate input. One instance is represented as a sequence of a concatenation of an instruction, input, and output, $[I;x;y]$, where instruction, input, and output are represented as textual sequences. Then, the instruction following task is formulated as a causal language modeling task by autoregressively predicting the next token in the sequence.  

\subsection{A Simplified Instruction-Following Task} \label{sec:syn_instruction}
For ease of analysis, we simplify the instruction-following task by making several assumptions. To emulate language data, we assume both alphabets $\mathcal{X}$ and $\mathcal{Y}$ are discrete. To have a focus of analysis, we assume the input and output are each represented by a single token. The next token prediction task allows us to have one token representation to evolve from the current token to the next token across the Transformer layers. We further assume the output token comes right after the input token without using any template tokens, allowing us to concentrate our study on representations of one single token across layers (i.e.\ its hidden states). 

To accommodate these assumptions, we synthesize a task function by randomly sampling a finite number of mappings such that an input of the function is uniquely associated with an element in $\mathcal{Y}$. The mapping could be made stochastic, so an input could be associated to multiple different output elements. For simplicity, we assume uniqueness. This reflects the fact that we usually only have a finite number of demonstrations for a task, for model learning and for evaluation. Different task functions may share an identical input set, but the respective outputs could be different, so it is important for the model to learn to correctly identify a task to provide accurate output accordingly. Additionally, our focus lies on the model's ability to generalize by identifying the correct task from unseen instructions, rather than the generalization of the task itself, which is beyond the scope of this study. Therefore, we provide all of the mappings from a task but only a portion of instructions to the model during the training process. 

\begin{figure*}[!htb]
\centering
\includegraphics[width=0.6\linewidth]{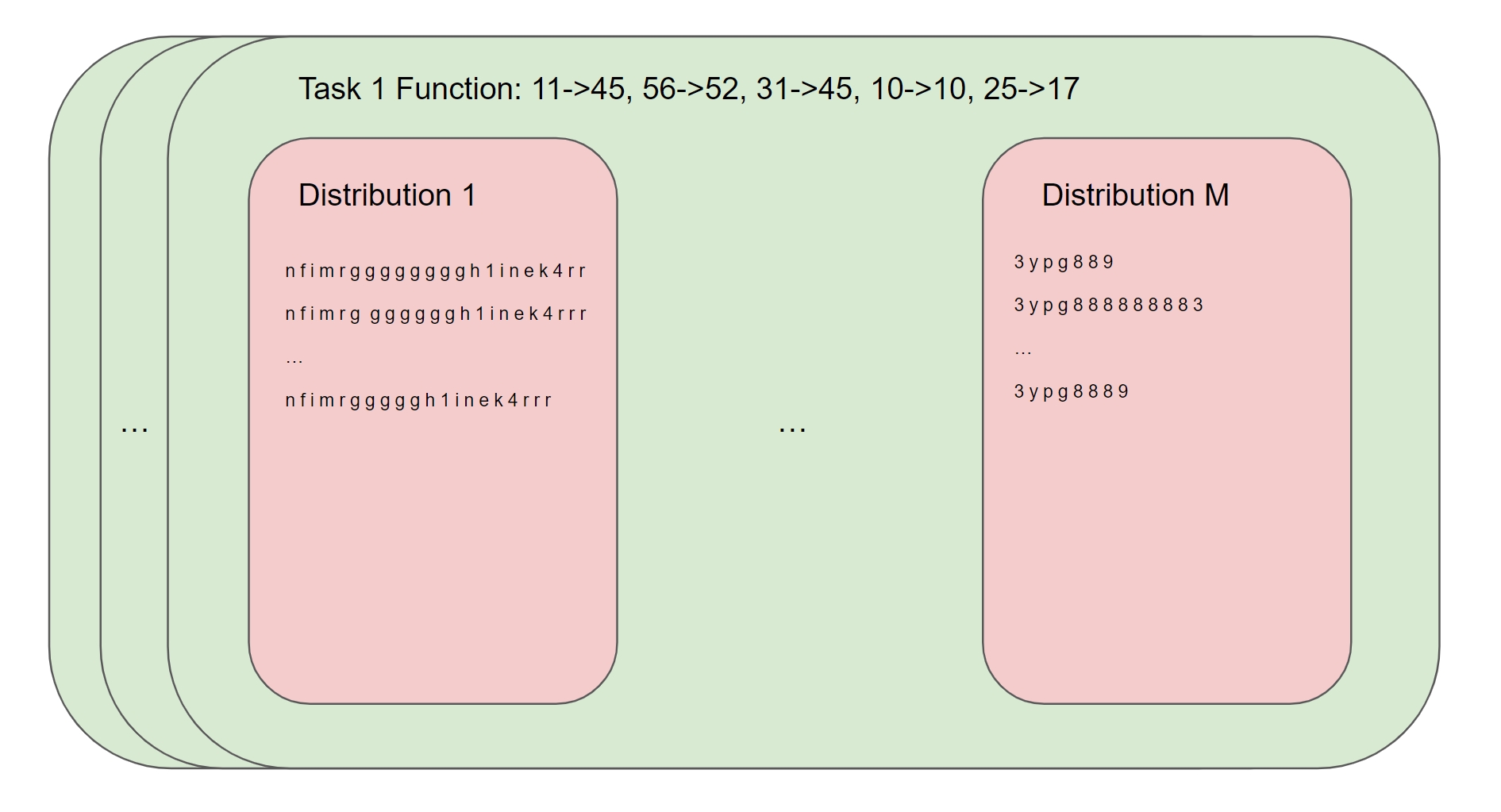}
\caption{A synthetic dataset for our simplified instruction-following setting. The first task colored green is shown as an example with instructions sampled from $M$ different distributions colored pink. The task function consists of five mappings, in which "->" means from an input to an output. There are several instructions sampled via a regular expression under each distribution.}
\label{fig:task_structure}
\end{figure*}

We aim to investigate the behavior of the model when provided with sufficient data to learn the patterns of different instructions. Given resource constraints, it is infeasible to create a large-scale instruction-following dataset of natural language that enables the model to fully comprehend the complexities inherent in natural language and then to train a Transformer model on such a vast dataset. Therefore, we opt to study the regularities of instructions simpler than those of natural languages using \emph{regular expressions}. We illustrate a synthetic dataset in Figure ~\ref{fig:task_structure}. Regular expressions are patterns used to match character combinations in strings. They are widely used in text processing and search tasks, allowing for flexible and powerful matching operations. We can also use regular expressions to synthesize as much data as needed by sampling based on these expressions, so the model can adequately acquire the ability to recognize regularities within the instructions. In our study, we randomly sample a regular expression, as detailed in Appendix~\ref{sec:appendix:si}. Each sampled regular expression is considered as a simple grammar rule. We then sample instructions represented as sequences of symbols based on the regular expression. We construct instructions by sampling instances based on different regular expressions to emulate different distributions. 

Another concern arises from many real-world tasks needing to acquire external knowledge. For instance, to learn to predict the next letter in the alphabet sequence, the model must possess external knowledge of the alphabet itself. Since we synthesize task functions in our approach, as outlined earlier, we can present all information to the model to learn how to solve a task, overcoming this limitation. We associate each task with instructions originating from distinct distributions and construct a data instance via concatenating an instruction and a mapping together. Each instruction will accordingly have a task identity. In this context, the different distributions highlight instructions characterized by highly distinct regularities, such as varying vocabularies and syntactic structures. Given the existence of task-specific clusters as shown in Figure~\ref{fig:learning_dynamics_f1}, this treatment also allows us to examine whether the model forms task-specific clusters based solely on the similarities of instructions. Moreover, to delve deeper into the Transformer model's ability to form task-specific clusters, we create hard examples by replacing a word within certain instructions in the training data with another word, thereby associating these instructions with a new task. For instance, in a realistic scenario, substituting the word ``initial'' with ``secondary'' from ``return the initial letter from the provided letter list'' indicates a different task. To further increase the difficulty, we introduce a new task with identical mappings as the original task but modify the outputs. In a realistic scenario, even a subtle change like this would likely trigger a different task. These hard examples can be viewed as outliers of the original data distribution. This creates instructions that are difficult to distinguish based solely on their appearance, posing a challenging task to assess whether the model can still effectively separate them into distinct clusters based on task identities.

\section{Experiments} \label{sec:experiment}
\subsection{Implementation Details} \label{sec:implementation_details}
We construct a synthetic instruction-following instruction dataset based on the guidelines outlined in Section~\ref{sec:syn_instruction}. This dataset is then divided into training and validation sets. For computational efficiency in subsequent clustering analysis, we randomly sample a number of instances from a subset of tasks to form the validation set and a training subset for intermediate evaluations. Given full control over the data generation process, we record meta information such as a task identity for each data instance. Further details regarding the hyperparameters of the data generation process and statistics of the resulting datasets are in Appendix~\ref{sec:appendix:si}. 

We train a six-layer Transformer model following the GPT-2 architecture~\citep{GPT-2}. This model is optimized using an AdamW optimizer~\cite{AdamW} and employs a cosine annealing learning rate schedule. We terminate training based on the best task accuracy achieved on the validation dataset. The task accuracy is measured by the percentage of correct outputs, which is treated as the measurement of task performance. Additional specifics about the hyperparameters of the model and training process are in Appendix~\ref{sec:appendix:hyperparameters}. We perform all of our experiments on a single NVIDIA A100 GPU. 

\subsection{Clustering Analysis} \label{sec:clustering_analysis}

\begin{figure*}[!htb]
\centering
\begin{subfigure}{\textwidth}\label{fig:syn_f1_trainsub}
  \centering
  \includegraphics[width=\linewidth]{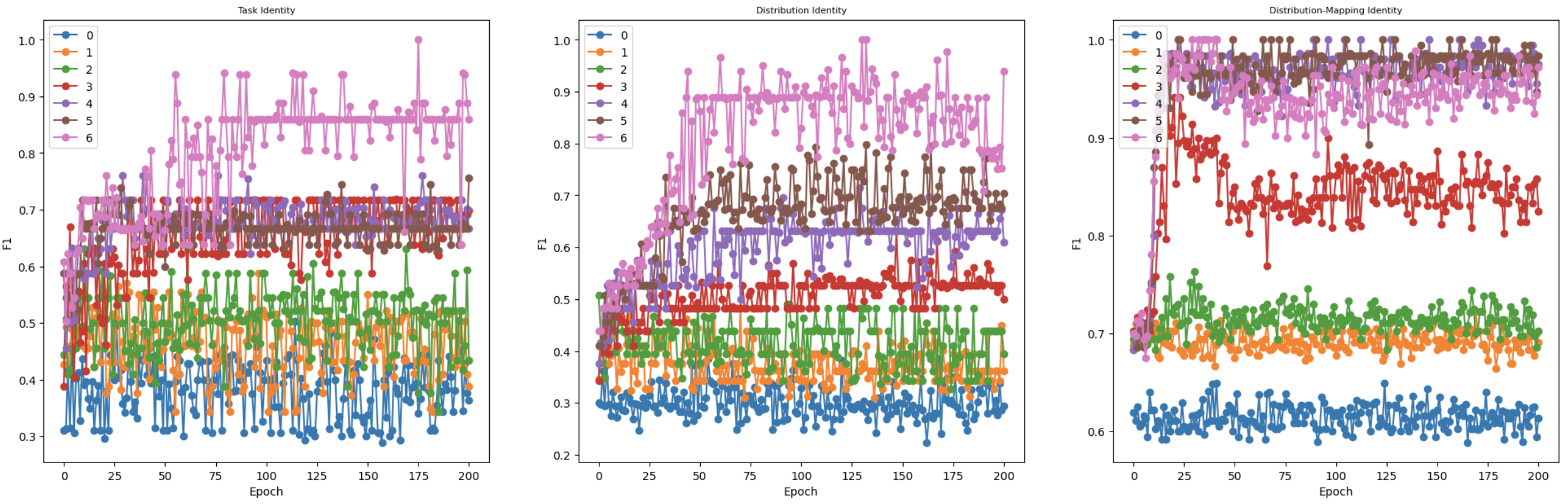}
  \caption{Training subset}
\end{subfigure}%
\\
\begin{subfigure}{\textwidth}\label{fig:syn_f1_val}
  \centering
  \includegraphics[width=\linewidth]{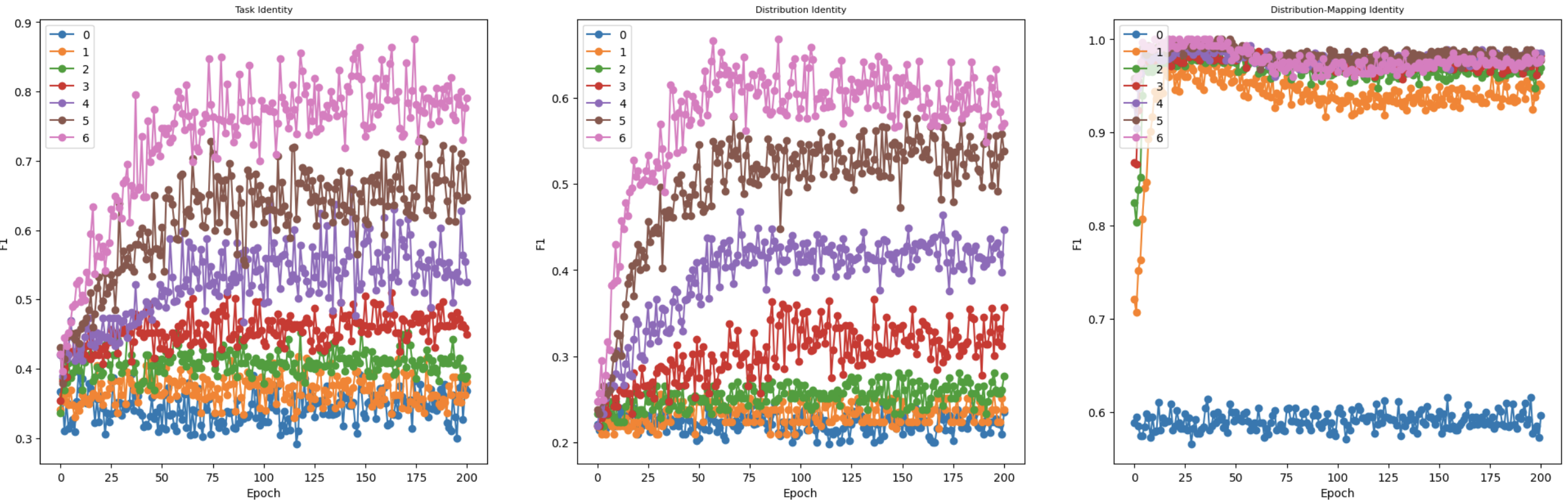}
  \caption{Validation set}
\end{subfigure}%
\caption{Clustering analysis on both of training subset (a) and validation set (b) across different layers throughout the training process: Different columns corresponds to uses of different identities as labels. Only shows results on F1 score here and see results on other evaluation metrics in Figure ~\ref{fig:learning_dynamics_all}. Each dot represents a data point.}
\label{fig:learning_dynamics_f1}
\end{figure*}

\begin{figure*}[!htb]
\centering
\begin{subfigure}{0.5\textwidth}
  \centering
  \includegraphics[width=\linewidth]{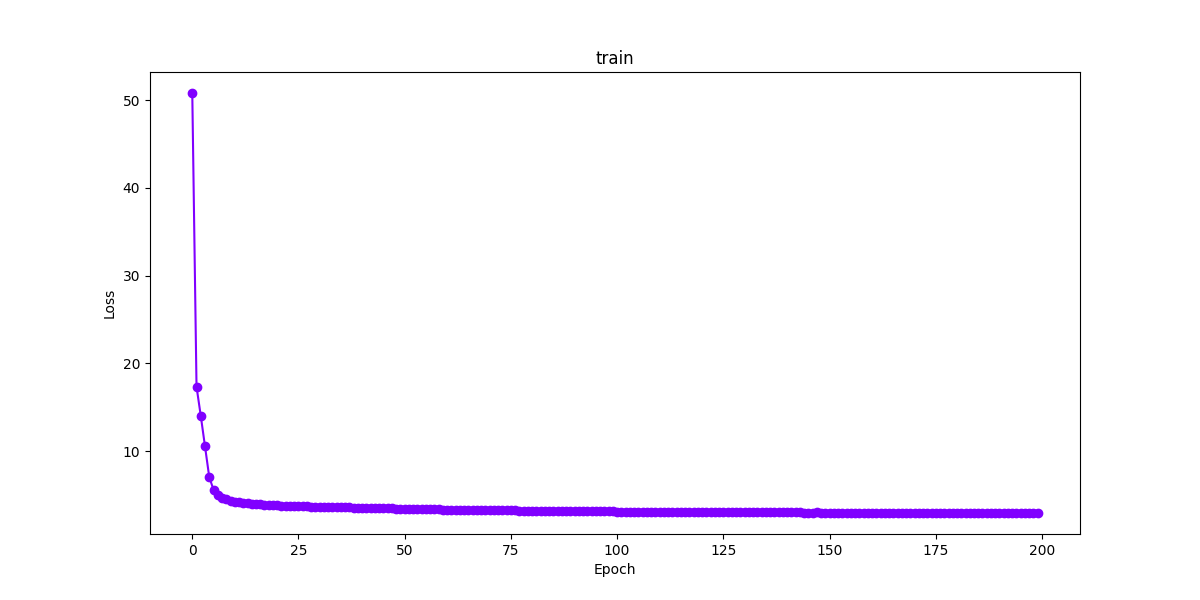}
  \caption{Training Loss}
  \label{fig:syn_training_loss}
\end{subfigure}%
\\
\begin{subfigure}{0.5\textwidth}
  \centering
  \includegraphics[width=\linewidth]{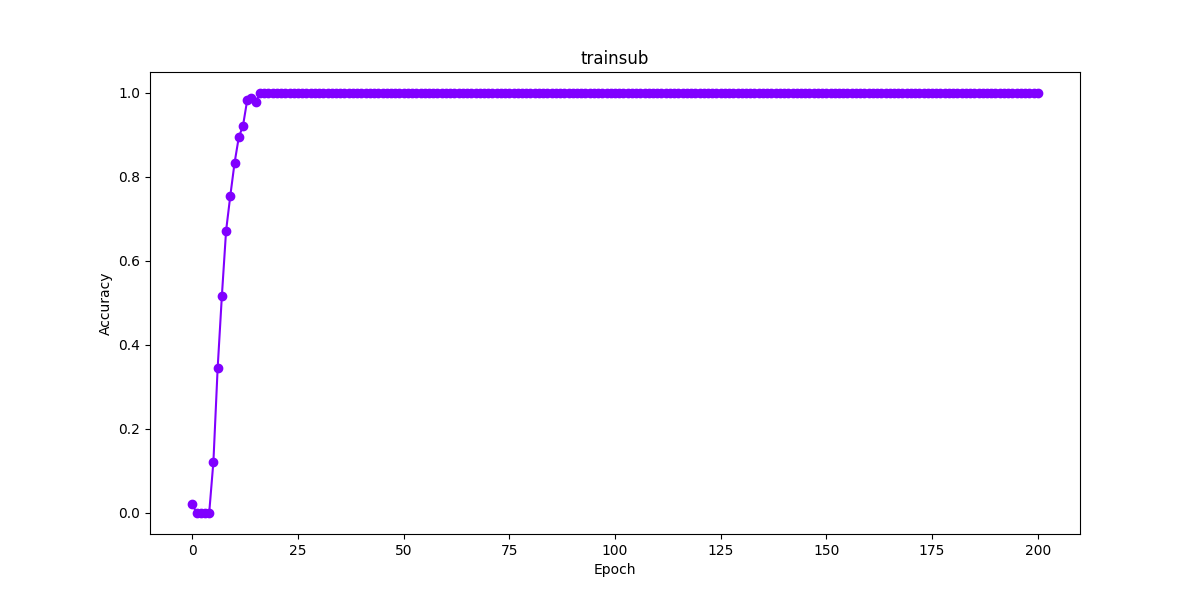}
  \caption{Trainsub Task Accuracy}
  \label{fig:syn_trainsub_acc}
\end{subfigure}%
\begin{subfigure}{0.5\textwidth}
  \centering
  \includegraphics[width=\linewidth]{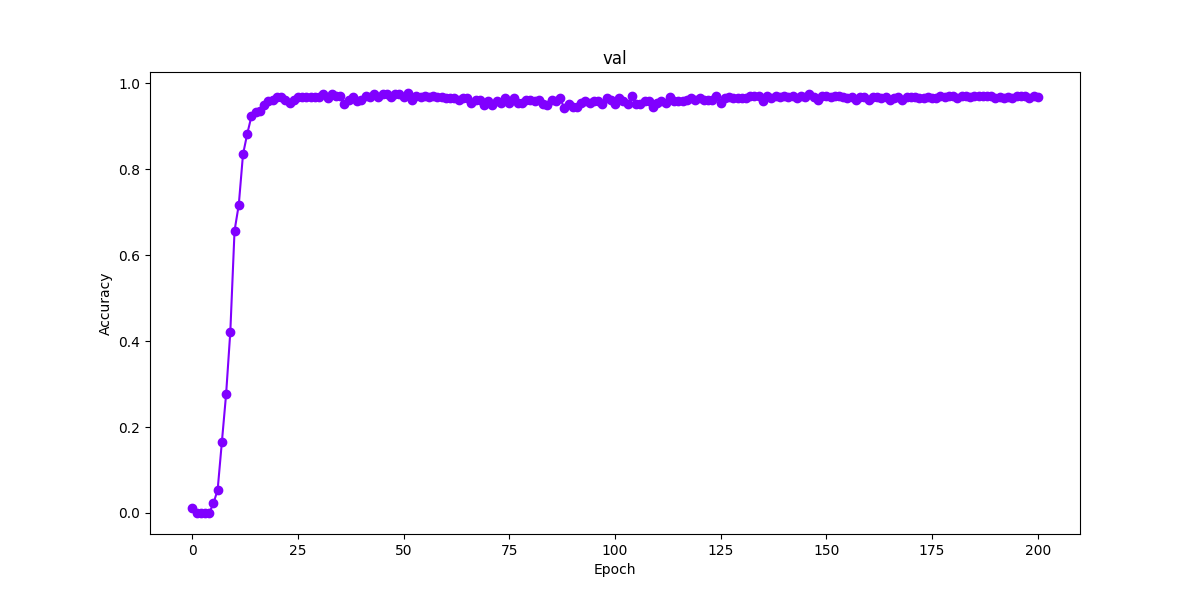}
  \caption{Validation Task Accuracy}
  \label{fig:syn_val_acc}
\end{subfigure}%
\caption{(a) Training loss, (b) Training subset task accuracy, and (c) validation task accuracy throughout the training process. Each dot represents a data point. Both (b) and (c) show dense dots with near-zero accuracy for the first few epochs.}
\label{fig:learning_curve}
\end{figure*}

As detailed in Section~\ref{sec:syn_instruction}, our study focuses on the hidden states corresponding to the input tokens (the last token in a sequence) within the Transformer model. We gather hidden states of the input tokens from various data instances. Next, we employ the popular KMeans clustering algorithm to uncover clusters within the data. We optionally pre-process them using t-SNE dimension reduction ~\citep{tsne} if it benefits the subsequent clustering performance. We conduct extrinsic clustering evaluation on the clustering results, utilizing task identities as labels. Our analysis reports results on the training subset and validation set, employing three commonly used metrics for clustering analysis: F1 score, adjusted rand index (ARI), and adjusted mutual information (AMI). Further details regarding the clustering analysis and evaluation metrics are in Appendix~\ref{sec:appendix:clustering_analysis}.

As shown in Figures \ref{fig:learning_dynamics_f1}, on both of training and validation splits, there exists a strong trend of improvement of the clustering performance based on task identities throughout the training process until saturating at some high values. Figure \ref{fig:learning_dynamics_all} in the Appendix demonstrates results on all evaluation metrics. Especially, during late stage of the training process, the hidden states exhibit a strong clustering effect on task identities, indicated by high values across different data splits and metrics. The model early stops at $51$th epoch, but particularly noteworthy is the persistence and even improvement of the clustering phenomenon long after the early stopping point, indicating clustering as a strong inductive bias of the Transformer during its training process. In addition, the early stop point is close to when the clustering performance starts to saturate. After that, the model's task performance also saturates at high values as shown in Figure ~\ref{fig:syn_val_acc}, which may indicate a correlation between the task performance and the clustering performance.  

Moreover, clustering performance tends to improve in higher layers of the Transformer model, with the $0$th layer serving as a baseline solely based on input word embedding. Notably, the baseline does not undergo much change during the training process compared to the clustering performances of other layers. It is important to note that task identities are concealed from the training process, and the Transformer models perform clustering during training without explicit supervision. Besides, we design the simplified task to have many tasks share the same inputs by using a small task-related vocabulary such that the model won't be able to identify a task solely from the inputs. 

Also, interestingly, based on our formulation of the instruction-following task in Section~\ref{sec:preliminary}, instances with instructions from different distributions are clustered together based on their task identities. This is corroborated by the high F1 score. Further, our analysis reveals that hard examples, formulated as described in Section~\ref{sec:syn_instruction}, and their corresponding original examples are predominantly separated into different clusters. This can be seen from the high F1 score of over 0.9 at the late training stage on the training subset, which contains both the hard examples and their original examples. This eliminates the possibility that the model groups instances solely based on instruction similarities. This underscores the model's ability to form clusters based on task identities. Additionally, similar clustering phenomena are observed on both the validation and testing sets, indicating that the clustering effect generalizes to unseen instances as well. These results not only provide compelling evidence supporting the existence of task-specific clusters but also show that the clusters evolve throughout the training process instead of appearing spontaneously.

We have found that task-specific clusters exist within the hidden representation space of the Transformer model. This suggests the question of whether there are any intriguing internal structures within these clusters. To explore this, we conduct the same clustering analysis using the distributions that the instructions belong to as labels. As demonstrated in Figures \ref{fig:learning_dynamics_f1}, we found obvious clustering effects based on this setting, which reveals possible inner clustering structures within the task-specific clusters. To delve deeper and uncover finer-grained structures within this hierarchical clustering, we conduct further analysis by considering a combination of labels, including the identities of instruction distributions and mappings. We observe strong clustering under this setting as well, as evidenced by high clustering performances.  See \cite{Maps2,Maps} for similar clustering and mapping phenomena in the mammalian cortex.

Figure~\ref{fig:learning_curve} showcases the learning curve of the model based on task accuracy. An intriguing observation is that the task accuracy remains around zero for the first few epochs on both training subsets and validation set before abruptly beginning to rise thereafter, despite continuous improvements in training loss as depicted in Figure~\ref{fig:syn_training_loss}. We hypothesize that the model initially learns task identification through the evolved clustering process by resolving various ambiguities introduced by us including the hard examples, enabling it to subsequently learn to solve different tasks successfully. We also confirm a similar clustering phenomenon and various trends we have discovered so far on a smaller model with a small hidden dimension of 32 and a larger model with a large hidden dimension of 2048. See results on this additional models in Figures ~\ref{fig:learning_dynamics_all_32}, ~\ref{fig:syn_KNN_32} ~\ref{fig:learning_dynamics_all_2048} and ~\ref{fig:syn_KNN_2048} of the appendix.

\subsection{Advantages of Clustering}

\begin{figure*}[!htb]
\centering
\begin{subfigure}{0.5\textwidth}
  \centering
  \includegraphics[width=0.8\linewidth]{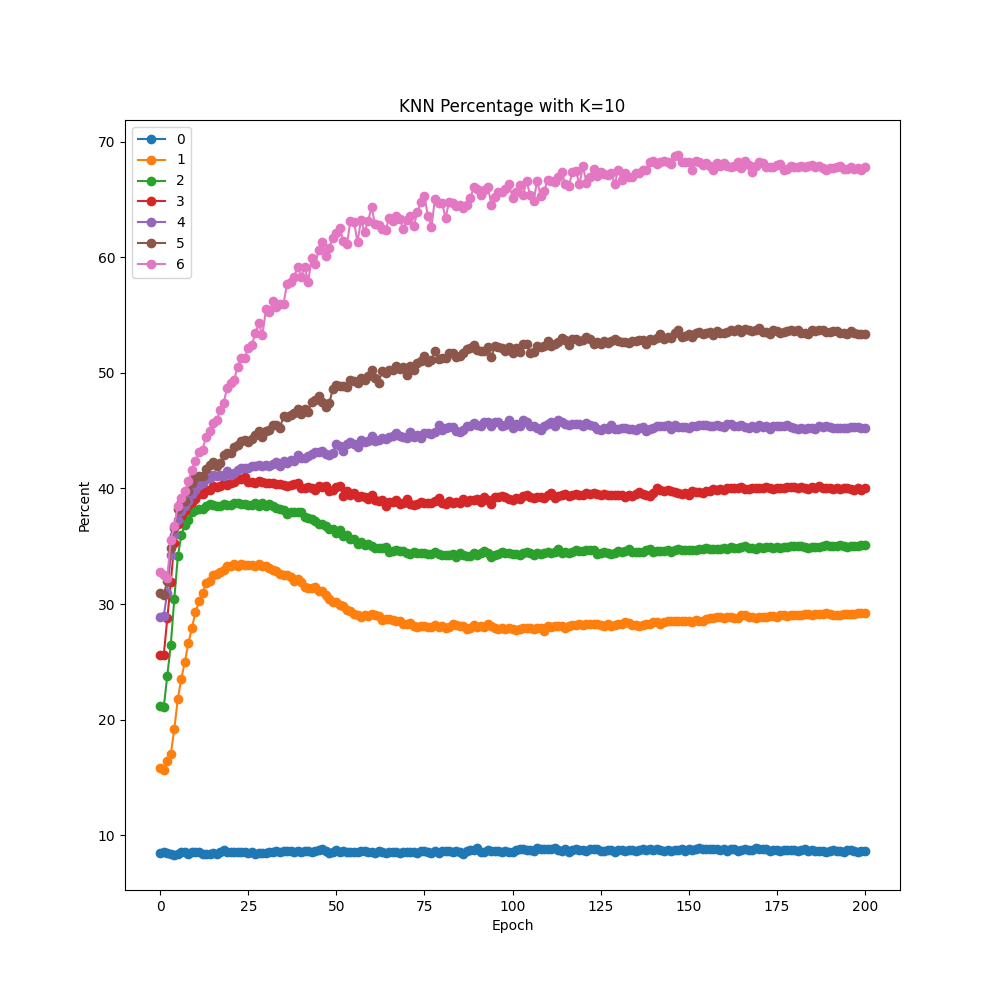}
  \caption{KNN Percentage}
  \label{fig:knn_task_10}
\end{subfigure}%
\begin{subfigure}{0.5\textwidth}
  \centering
  \includegraphics[width=0.8\linewidth]{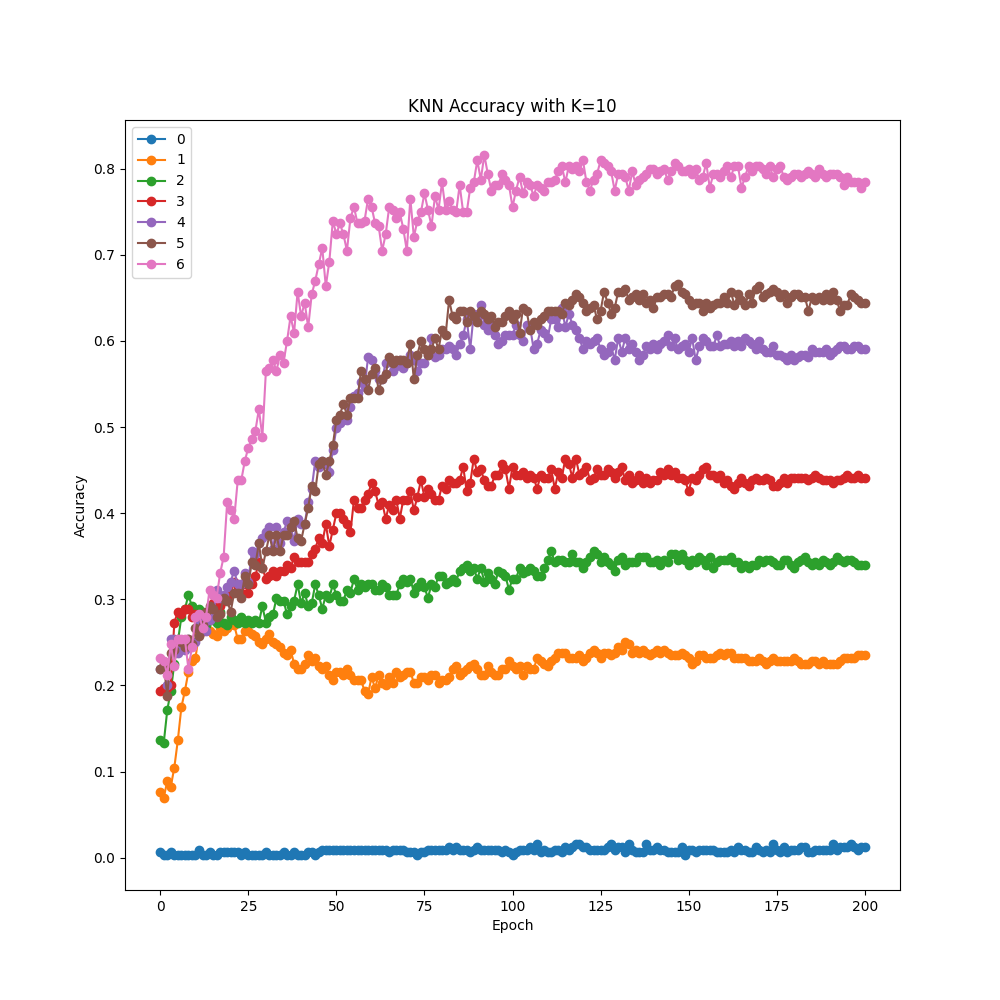}
  \caption{KNN Accuracy}
  \label{fig:knn_acc_10}
\end{subfigure}%
\caption{(a) Percentage of K nearest neighbors in the training set of an unseen instance belonging to the same task identity. (b) K nearest neighbors accuracy. Measurements are performed across all layers and throughout the training process.}
\label{fig:syn_KNN}
\end{figure*}

Next, we explore the potential advantages of the clustering phenomenon observed earlier. We have observed clustering effects on both the training subset and unseen instances from the validation set. To further verify if training instances and unseen instances with the same task identity are close in the hidden representation space, Figure~\ref{fig:knn_task_10} shows the percentage of $K$ nearest training instances of an unseen instance belonging to the same task identity, averaged over all instances. We observe a dramatic improvement in the percentage along the training process, indicating that both training instances and unseen instances are not only close in the hidden space but also become more clustered as the training proceeds. This suggests that the same task-specific clustering structure generalizes to the unseen instances.

Previous work~\citep{KNN-LLM} has demonstrated that using an inference method based on K-Nearest Neighbors (KNN) algorithms with pre-trained Transformer-based language models can achieve competitive or even better next token prediction performance than inference methods based on models' forward pass. This inspires us to record the task performance of our models based on the KNN during the training process. From Figure~\ref{fig:knn_acc_10}, we observe that the task accuracy based on KNN improves consistently during training until saturating at high values, providing direct evidence of the advantages of task-specific clustering by bringing instances of the same task closer together in the hidden space. More specifically, the model clusters instances belonging to the same task close to each other such that it is easy for making inferences for even unseen instances by using their nearby data instances. The KNN accuracy is also improved across layers. This is also apparently a working way to identify a specific task by a model gradually moving a representation to those with the same task over a series of layers.  

\subsection{Analysis of Natural Instruction-Following Task} \label{sec:natural_instruction}

\begin{table*}
\centering
\begin{tabular}{llllllllll}
\hline
\multicolumn{1}{c}{\bf Model} &\multicolumn{3}{c}{\bf Task} &\multicolumn{3}{c}{\bf Distribution} &\multicolumn{3}{c}{\bf Distribution-Mapping}\\
 & \textbf{F1} & \textbf{ARI} & \textbf{AMI} & \textbf{F1} & \textbf{ARI} & \textbf{AMI} & \textbf{F1} & \textbf{ARI} & \textbf{AMI}\\
\hline
LLaMa-7B & 0.959 & 0.872 & 0.869 & 0.571 & 0.438 & 0.673 & 0.940 & 0.903 & 0.974\\
LLaMa-13B & 0.953 & 0.855 & 0.854 & 0.546 & 0.381 & 0.645 &  0.936 & 0.893 & 0.969\\
GPT-J-6B & 0.887 & 0.697 & 0.665 & 0.652 & 0.485 & 0.653 & 0.465 & 0.345 & 0.602 \\
LLaMa-2-7B-Instruct & 0.917 & 0.756 & 0.780 & 0.563 & 0.339 & 0.638 & 0.932 & 0.890 & 0.966 \\
Instruct-GPT-J-6B &0.907 & 0.749 & 0.692 & 0.471 & 0.290 & 0.516 & 0.170 & 0.084 & 0.342 \\
\hline
\end{tabular}
\caption{Clustering analysis on open LLMs using task identity, distribution identity, and distribution-mapping identity as labels.}
\label{tab:LLMs_clustering}
\end{table*}

One further question to ask is if the clustering phenomenon we discovered under the simplified setting generalizes to realistic settings. Therefore, we studied trained LLMs on a realistic instruction-following task based on natural language to supplement our analysis. We utilize tasks and their descriptions in natural language from~\citep{task_vector} of three categories: Knowledge, Linguistic, and Translation. In accordance with the typical approach of constructing instruction datasets via LLM self-instruct~\citep{self_instruct}, we build a set of instructions for each task by using their task descriptions as seeds to prompt ChatGPT~\citep{chatGPT} to generate 50 different expressions. We consider expressions sampled based on the same seed as coming from the same distribution. The specifics of the task descriptions and prompts used for querying ChatGPT are in Appendix~\ref{sec:appendix:natural_if}. The subsequent step involves linking each instruction to a task mapping provided by \citep{task_vector} in the same way as described in Section~\ref{sec:syn_instruction}. We only keep those task mappings that have inputs and outputs of only a single word to have better focus of study. For computational efficiency, we only use ten of those selected task mappings for each task. We assign the same task identity to tasks under the same category due to their similarities. Actually, the instruction-following data can be considered as a spacial kind of language data and naturally exist in the large-scale language data used for pre-training. Therefore, we will perform the same clustering analysis as in Section~\ref{sec:clustering_analysis} on several different open LLMs either instruction tuned or not: LLaMa-7B, LLaMa-13B~\citep{llama}, GPT-J-6B ~\citep{gpt-J-6B}, LLaMa-2-7B-Instruct~\citep{llama2,Llama-2-7B-32K} and Instruct-GPT-J-6B~\citep{instruct-gpt-j-fp16,alpaca_data}, in which LLaMa-2-7B-Instruct and Instruct-GPT-J-6B are fine-tuned on instruction datasets. B (billion) refers to the number of parameters. See more details about the model sizes in Table ~\ref{table:model_size} in the appendix. 

We only report measurements of the layer with the best F1 score on clustering based on task identity. As shown in Table~\ref{tab:LLMs_clustering}, similar to our results in the simplified setting, all of the LLMs achieve high clustering performances based on task identities. In particular, a high F1 score indicates different tasks under the same category are clustered together. Both the LLaMa models of different sizes and the LLaMa-2-7B-Instruct models receive high scores on clustering instances with the same distribution-mapping identities, which is consistent with our results on the simplified setting. However, the GPT-J-6B and Instruct-GPT-J-6B seem to not form clear clusters based on the distribution-mapping identity. Besides, as we expected, the clustering phenomenon appears on LLMs either instruction tuned or not. We should note that the conclusions we made on the simplified setting may not completely extend to the realistic settings due to differences in data complexity and scales. However, we can still see some consistent results, which indicates some universality of the clustering phenomenon. 
We hope our analysis of both the simplified and realistic settings can shed some light on understanding the inductive biases of the Transformer-based LLMs for the instruction following. 

\section{Applications}\label{sec:applications}
In this section, we will demonstrate two applications under our simplified setting inspired by our findings: pre-training a model using task identities and an alignment algorithm with less forgetting. 

\subsection{Pre-training with Task Identities}\label{sec:task_id_pretraining}
Our experimental results from Figure~\ref{fig:learning_dynamics_f1} present task-specific clusters evolved from the training process for the instruction-following task. This inspires us to ask if we can directly guide a model to learn the clustering structures within its hidden representation space before being tuned for various tasks so the tuning process becomes more efficient. 

\begin{figure*}[!htb]
\centering
\begin{subfigure}{0.5\textwidth}
  \centering
  \includegraphics[width=0.8\linewidth]{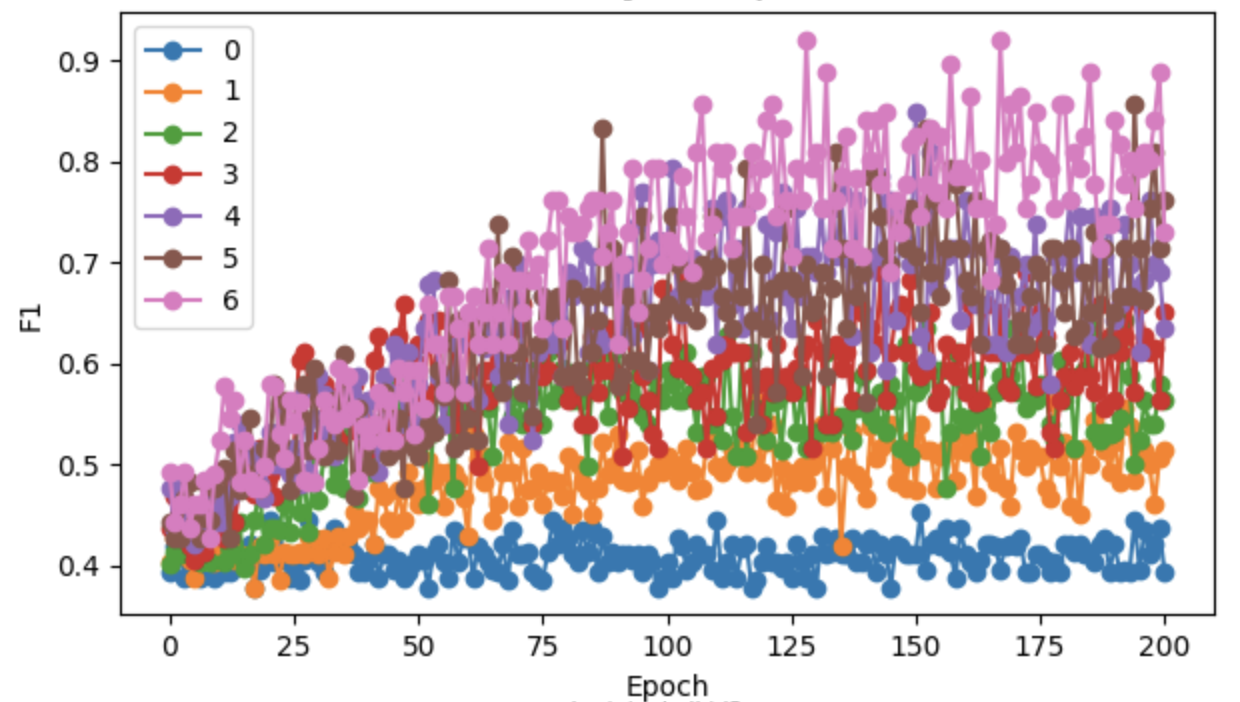}
  \caption{}
  \label{fig:taskid_clustering}
\end{subfigure}%
\begin{subfigure}{0.5\textwidth}
  \centering
  \includegraphics[width=0.8\linewidth]{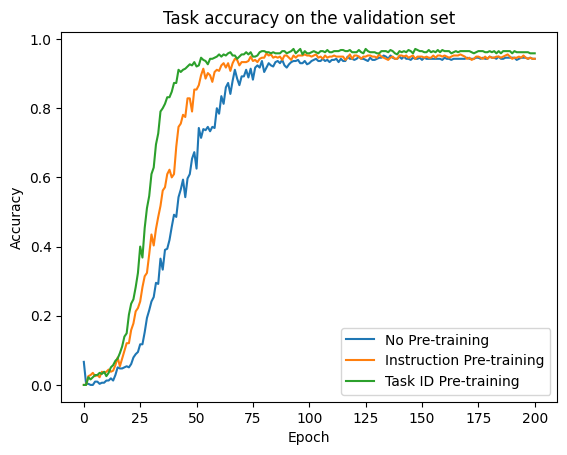}
  \caption{}
  \label{fig:taskid_comparison}
\end{subfigure}%
\caption{(a) Clustering analysis of the model trained for the task ID prediction following the setting in Section ~\ref{sec:clustering_analysis}. (b) Comparison of different pre-training strategies by their performances during the fine-tuning process. The task accuracy is measured on the validation set.}
\label{fig:task_id_pretraining}
\end{figure*}

To verify this idea, we pre-train a CLM model to predict a task identity (ID) for a given instruction. More specifically, we append a special token to the end of an instruction to trigger the model to predict the corresponding task ID as the next token. We select the model with the best prediction performance on the validation set for fine-tuning at the next stage. Please note that the model is pre-trained by a regular causal language modeling task over an entire sequence instead of solely the task ID prediction task. Also, the task mappings are not used at the pre-training stage. As shown in Figure~\ref{fig:taskid_clustering}, the task-specific clusters indeed evolve during the pre-training process. After the pre-training, we fine-tune the model for the instruction-following task and measure the task accuracy over the fine-tuning process. We compare this pre-training strategy with direct training for the instruction-following task denoted by \emph{No Pre-training}. 

To see if our pre-training approach makes additional contributions besides language modeling over the instructions, we also compare to a setting (Instruction Pre-training) that pre-trains a model by causal language modeling over only the instructions. A model for the following fine-tuning is chosen based on the minimum validation CLM loss. We focus on a smaller model with a hidden dimension of 256 since there are two stages of training.

From Figure~\ref{fig:taskid_comparison}, we see that our method not only converges faster but also achieves higher task accuracy on the validation set during the fine-tuning process than the other two settings. The comparison to the Instruction Pre-training setting indicates the information about task ID benefits the downstream fine-tuning. The results demonstrate both the efficiency and effectiveness of our pre-training method. It may be feasible to perform this kind of pre-training in practice since information about task identities is not hard to collect. Further, due to the high cost of LLM fine-tuning, shortening its training schedule by accelerating its convergence is both economically and environmentally appealing.  

\subsection{Alignment}\label{sec:alignment}
One goal in alignment is to reduce undesirable behaviors of LLMs such as toxicity via additional tuning. Aligning a model to reduce toxicity can be simulated in our simplified setting: We assume there is a toxic behavior represented as a task triggered by some instructions with toxic tendencies. We want to convert the toxic source task to a healthier target task that is associated with instructions of goodwill and shares the same input domain. Then, a model gives outputs corresponding to the healthier task, even if toxic instructions are seen. We consider the model to get exposure to the healthier task during its training because this is close to the reality, in which the model is trained on large-scale language data. 

We first train a model on an instruction-following dataset generated according to our simplified setting, in which we label some tasks as toxic, some as their healthier counterparts, and others as regular tasks. Next, we construct a new and smaller dataset consisting of only the data relevant to the toxic tasks by updating the outputs of the toxic tasks according to the target tasks and keeping the toxic instructions unchanged. We fine-tune the model using the new dataset to perform the alignment. 

A straightforward idea is direct fine-tuning, but this may lead to unwanted catastrophic forgetting of other tasks \citep{forgetting}. To overcome this drawback, one can fine-tune only 
the linear language head by freezing the rest of the model. This approach could be useful in practice since we are only expected to make limited changes to the model's behavior as simple as updating outputs for certain tasks. However, updating the language head may still impact the performance of other tasks because all tasks share the same language head. Since clusters found by Kmeans may have good separability, we are inspired to use a switch network that activates a new language head for the source tasks and the original language head otherwise based on the hidden states of the last tokens from the last layer. Then, only the new language head and the switch network are trained during the fine-tuning process such that the performance of other tasks could largely remain. Since the clustering structure is clear according to our experiments in Section~\ref{sec:clustering_analysis}, we use instances of the source tasks and only one instance for each other task to train the switch network to classify whether an instance belongs to a source task or not. 

We consider both a linear switch network implemented by a linear layer and a switch network constructed by a more capable three-layer multilayer perceptron (MLP). The intermediate dimension of the MLP is the same as the hidden dimension of the CLM model. We compare different methods by selecting a model based on the best average validation accuracy of updated source tasks.

\begin{table*}
\centering
\begin{tabular}{lll}
\hline
\multicolumn{1}{c}{\bf Method} &\multicolumn{1}{c}{\bf Task Accuracy} &\multicolumn{1}{c}{\bf Accuracy Drop}\\
\hline
Direct fine-tuning & \bf 1.00 & 0.210 \\
Language head fine-tuning & \bf 1.00 & 0.120 \\
Ours with Linear switch network & \bf 1.00 & 0.087 \\
Ours with MLP-based switch network & \bf 1.00 & \bf 0.080 \\
\hline
\end{tabular}
\caption{Comparison of different alignment methods described in Section ~\ref{sec:alignment} by their performances on a validation set. Task accuracy is measured for updated source tasks only and accuracy drop is measured for tasks other than the source tasks.}
\label{tab:alignment}
\centering
\end{table*}

Table~\ref{tab:alignment} shows that all methods achieve perfect performance for the updated source tasks on a validation set, but the direct fine-tuning undergoes significant performance drops on other tasks. Note that perfect performance means a model provides updated and healthier outputs even if toxic instructions are given. The method that only fine-tunes the language head retains more performance on other tasks, but not as much as methods using the linear switch network and the MLP-based switch network. The MLP-based switch network is slightly better than the linear switch network. Moreover, a few additional examples of the other tasks are not only attainable in practice but also are sufficient to achieve the small performance degradation, which further supports the good separability. The performance of other tasks is not completely preserved due to the imperfection of the switch network and more examples may be used to obtain a better switch network.  

More experimental details are in Appendix~\ref{sec:appendix:hyperparameters}. The two applications could be further verified in a more realistic settings, but is left as future work due to limited computational resources and desire to constrain the the scope of this work.  

\section{Related Work}

\subsection{Instruction Following}
Making LLMs follow user intention specified in instructions is important for making them more truthful and less toxic. Many efforts has been made to achieve this goal \citep{hill2020human,IT_survey,RLHF,DPO}. In our work, we focus on studying the hidden mechanism and inductive biases of the Transformer-based CLMs to learn instruction following in a general and simplified setting, rather than developing a more advanced instruction-following method.  

\subsection{Functional Vectors}

Recent works \citep{function_vector,task_vector} present compelling evidence of function vectors that store task-related information in in-context learning. While in-context examples can be viewed as a specific type of instruction, our work primarily focuses on conducting analysis based on more general textual instructions rather than in-context examples. Further, our study extends to analyzing the learning dynamics of models, rather than solely focusing on trained models.

\subsection{Mechanistic Interpretability}
The primary objective of mechanistic interpretability is to reverse engineer model behaviors \citep{olah2020zoom, elhage2021mathematical, nanda2023emergent, meng2022locating, hernandez2023linearity, geva2023dissecting, conneau2018you, ilharco2022editing}. Similar to many of the works in this area, we conduct studies based on a synthetic task and data to gain better controllability of the experiments and perform more in-depth analysis. 

\subsection{Clustering in Transformers}
Some studies also explore clustering phenomena within the Transformer model ~\citep{chen2021probing,reif2019visualizing,geshkovski2023mathematical,thompson2020topic}. However, they did not specifically focus on the instruction-following setting and conducted analysis mainly on trained models. \citet{geshkovski2023mathematical} primarily concentrates on studying clustering among tokens within a sequence. In contrast, our clustering analysis focuses on identifying clustering structures among different sequences.

\section{Conclusion}
In this work, we introduce a simplified instruction-following task and construct synthetic datasets to analyze a Transformer-based CLM model. From the simplified setting, we provide experimental evidence supporting the notion that the model encodes task-specific information through clustering in its hidden space, and demonstrate that this clustering evolves continuously during the learning process. Additionally, we highlight the advantages of the clustering phenomenon for the model to handle unseen instances. We also further verify the existence of the clustering phenomenon we discovered from the simplified setting in a realistic setting. The inductive biases uncovered and analyzed in this study offer new insights into Transformer-based CLM models and shed light on their remarkable instruction-following capabilities. Furthermore, this newfound understanding can inspire the development of more advanced algorithms to enhance LLM's capability to effectively follow human instructions. We have shown two inspired applications regarding pre-training and alignment respectively in this work. We will continue to explore more possibilities in the future.

\section{Limitations}
At present, our study is confined to a simplified task and synthetic dataset along with specific data distributional assumptions. Expanding the analysis to encompass a broader range of diverse and realistic distributional assumptions on data is an avenue for future exploration. We anticipate that our study can provide insights into Transformer's hidden mechanisms and inductive biases, serving as a foundational starting point and offering directions for analysis on larger scales. It is conceivable that our findings may have broader applicability and could be validated across a wider array of scenarios beyond our simplified instruction-following tasks. Scaling up our analysis to encompass more complex and realistic scenarios and verifying the inspired applications in a more realistic setting are areas we plan to explore in future research endeavors.

\bibliography{custom}

\appendix

\section{Simplified Setting}
\label{sec:appendix:si}

To construct data for the synthetic instruction-following task under the simplified setting, we first sample a task function consisting of several unique mappings. For each mapping, we sample two symbols from a task symbol vocabulary to form a mapping. We enforce that no two functions share a mapping. Next, we sample instructions for each task based on a regular expression. Regular expressions are sequences of characters that define a search pattern. They are widely used in computing for tasks such as text processing, string manipulation, and pattern matching. Regular expressions consist of normal characters (like letters and digits) and special characters (also known as metacharacters) that have special meanings. These metacharacters allow you to specify rules and conditions for matching patterns within text. We use regular expression reversely by sampling a string from a search pattern. To sample a regular expression, we first sample several metacharacters and then sample normal characters from an instruction vocabulary as their arguments and concatenate them together as a regular expression. For computational efficiency, we build a training subset, validation set, and hard examples from a subset of tasks by randomly selecting several instructions sampled from all of the distributions associated with each of the tasks. Please refer to Table ~\ref{table:data_hyperparameters} for related hyperparameters. We emulate sampling from different distributions by sampling from different regular expressions as described in Section \ref{sec:syn_instruction}.

We present the statistics of the sythetic instruction-following dataset in Table ~\ref{table:data_statistics}.

\begin{table*}
\centering
\begin{tabular}{lll}
\hline
\textbf{Setting}  & \textbf{Set} & \textbf{Size} \\
\hline
Simplified & Training & 7,300\\
 & Training subset & 180\\
 & Validation & 315\\
 Realistic & Testing & 8,800\\
\hline
\end{tabular}
\caption{\label{table:data_statistics}
Data Statistics of both simplified and realistic settings. The size of a data set is quantified by its number of instances. Only testing set is available for the realistic setting since we use pre-trained models instead of we training and validating a model in this setting. 
}
\end{table*}

\section{More on Clustering Analysis}
\label{sec:appendix:clustering_analysis}
In this work, we use extrinsic evaluation for our clustering analysis. Extrinsic evaluation of clustering refers to assessing the quality of clustering results by comparing them to ground truth. Ground truth data refers to labeled data that indicates the class or cluster to which each data point belongs. We utilize three widely used evaluation metrics: F1, adjusted rand index, and adjusted mutual information. 

{\bf F1 Score:} The F1 score combines both precision and recall into a single value, making it a useful measure of a model's accuracy. The formula for the F1 score is \(2 \times \frac{{\text{precision} \times \text{recall}}}{{\text{precision} + \text{recall}}}\). The F1 score ranges from 0 to 1 with higher values indicating better agreement to the ground truth. 

{\bf Adjusted Rand Index (ARI):} ARI is a measure of the similarity between two clustering results. It considers all pairs of samples and counts pairs that are assigned to the same or different clusters in both the true and predicted clusterings. ARI ranges from -1 to 1, where 1 indicates perfect clustering agreement, 0 indicates clustering results are random, and negative values indicate less agreement than expected by chance.

{\bf Adjusted Mutual Information (AMI):} AMI is another measure used to evaluate the quality of clustering. It quantifies the amount of information obtained about one clustering from knowing the other, adjusting for chance. Like ARI, AMI ranges from -1 to 1, where higher values indicate better agreement between clusterings.

\section{Hyperparameters}
\label{sec:appendix:hyperparameters}
Tables ~\ref{table:data_hyperparameters} and ~\ref{table:alignment_data_hyperparameters} show the hyperparameters of the data generation processes. Tables ~\ref{table:model_hyperparameters} contain the hyperparameters of our Transformer model and its training process. T-SNE related hyperparameters are listed in Table ~\ref{table:tsne_hyperparameters}

\begin{table*}
\centering
\begin{tabular}{ll}
\hline
\textbf{Hyperparameter} & \textbf{Value} \\
\hline
Number of tasks & 50 \\
Maximum number of instruction distributions per task & 6 \\
Minimum number of instruction distributions per task & 1\\
Number of instructions per distribution & 10 \\
Number of mappings per task & 5 \\
Number of tasks in training subset & 5\\
Number of instructions per distribution in the training subset & all available\\
Number of tasks in the validation set & 10\\
Number of instructions per distribution in the validation set & 3\\
Number of different tasks in hard examples & 5\\
Number of instructions per distribution in hard examples & 3\\
Size of the task symbol vocabulary & 25\\
Size of the instruction symbol vocabulary & 35\\
Maximum number of metacharacters per regular expression & 3 \\
Minimum number of metacharacters per regular expression & 1 \\
Maximum number of characters per metacharacters & 10 \\
Minimum number of characters per metacharacters & 3 \\
\hline
\end{tabular}
\caption{\label{table:data_hyperparameters}
Hyperparameters used for the data generation process.
}
\end{table*}

\begin{table*}
\centering
\begin{tabular}{ll}
\hline
\textbf{Hyperparameter} & \textbf{Value} \\
\hline
Learning rate & 1E-4 \\
Number of epochs & 200 \\
Optimizer & AdamW \\
Max gradient normM & 1.0 \\
validation criterion & Task accuracy \\
Scheduler & Cosine Annealing \\
Number of layers & 6 \\
Number of heads & 8 \\
Hidden dimension & 768 \\
feedforwark network dimension & 1024 \\
droptout & 0.2 \\
\hline
\end{tabular}
\caption{\label{table:model_hyperparameters}
Hyperparameters related to our model in the main experiment and its training. 
}
\end{table*}

\begin{table*}
\centering
\begin{tabular}{ll}
\hline
\textbf{Hyperparameter} & \textbf{Value} \\
\hline
Number of Components & 3 \\
Perplexity & 10 \\
Number of iterations & 2,000 \\
Metric & Euclidean \\
Initialization method & PCA \\
\hline
\end{tabular}
\caption{\label{table:tsne_hyperparameters}
T-SNE Hyperparameters ~\citep{tsne}. 
}
\end{table*}

\begin{table*}
\centering
\begin{tabular}{ll}
\hline
\textbf{Hyperparameter} & \textbf{Value} \\
\hline
Number of tasks & 50 \\
Number of source-target pairs & 3 \\
Maximum number of instruction distributions per task & 6 \\
Minimum number of instruction distributions per task & 1\\
Number of mappings per task & 5 \\
Number of tasks in the validation set & 10\\
Number of instructions per distribution in the validation set & 3\\
Size of the task symbol vocabulary & 65\\
Size of the instruction symbol vocabulary & 35\\
Maximum number of metacharacters per regular expression & 3 \\
Minimum number of metacharacters per regular expression & 1 \\
Maximum number of characters per metacharacters & 10 \\
Minimum number of characters per metacharacters & 3 \\
\hline
\end{tabular}
\caption{\label{table:alignment_data_hyperparameters}
Hyperparameters of the data generation process for the alignment experiments in Section ~\ref{sec:alignment}. We don't consider hard examples in this setting.
}
\end{table*}

\begin{table*}
\centering
\begin{tabular}{lll}
\hline
\textbf{Model} & \textbf{Hidden Dimension} & \textbf{Parameter Count}\\
\hline
Our model 768 & 768 &  23 million\\
Our model 32 & 32 &  55 thousand\\
Our model 256 & 256 &  3 million\\
Our model 2048 & 2048 &  202 million\\
LLaMa-7B & 4096 & 7 billion \\
LLaMa-13B & 5120 & 13 billion \\
LLaMa-2-7B-instruct & 4096 & 7 billion \\
GPT-J-6B & 4096 & 6 billion \\
Instruct-GPT-J-6B & 4096 & 6 billion\\
\hline
\end{tabular}

\caption{\label{table:model_size}
Sizes of models used in this work in terms of parameter counts and size of hidden dimension. The names of our models trained in the simplified setting end with their hidden dimension sizes.
}
\end{table*}

\section{Natural Instruction-Following Task}
\label{sec:appendix:natural_if}

\subsection{ChatGPT Prompt Template}
We use the following prompt template to query ChatGPT to generate different expressions of a task description: "Rewrite 50 different expressions of XXX", where "XXX" is a task description.   

\subsection{Realistic Setting}
See Table ~\ref{table:task_descriptions} for the task descriptions used for constructing the dataset for the realistic setting as detailed in Section ~\ref{sec:natural_instruction} and data statistics in Table ~\ref{table:data_statistics}.

\label{sec:appendix:task_descriptions}
\begin{table*}
\centering
\begin{tabular}{lll}
\hline
\textbf{Category} & \textbf{Task} & \textbf{Description}\\
\hline
Translation & French to English & Given a word in French, translate to English \\
 & English to French & Given a word in English, translate to French \\
 & Spanish to English & Given a word in Spanish, translate to English  \\
 & English to Spanish & Given a word in English, translate to Spanish \\
 & Italian to English & Given a word in Italian, translate to English  \\
 & English to Italian & Given a word in English, translate to Italian \\
Linguistic & Antonyms & Given an English adjective, output an antonym\\
 & plural to Singular   & Given an English noun in plural form, output the singular form \\
& Singular to plural   & Given an English noun in singular form, output the plural form \\
& Present to gerund & Given an English verb in present simple tense,\\
& & output the corresponding gerund form \\
 & Present to past perfect & Given an English verb in present simple tense, output the \\ 
 & & corresponding verb in past perfect \\
 & Present to past simple & Given an English verb in present simple tense, \\
 & & output the corresponding verb in past simple \\
 Knowledge & Country to Capital & Given a name of a country, output the name of the capital city \\
& Location to continent & Given a name of a location, output the name of its continent \\
& Religion & Given a name of a location or a person, \\
& & output the associated religion\\
 & Person to Language & Given a name of a person, output their native language \\
\hline
\end{tabular}
\caption{\label{table:task_descriptions}
Task descriptions provided by ~\citep{task_vector}}
\end{table*}

\section{More Results}
We present results obtained on various models here. 

\label{sec:appendix:more_results}

\begin{figure*}[!htb]
\centering
\begin{subfigure}{\textwidth}\label{fig:syn_all_trainsub}
  \centering
  \includegraphics[width=0.7\linewidth]{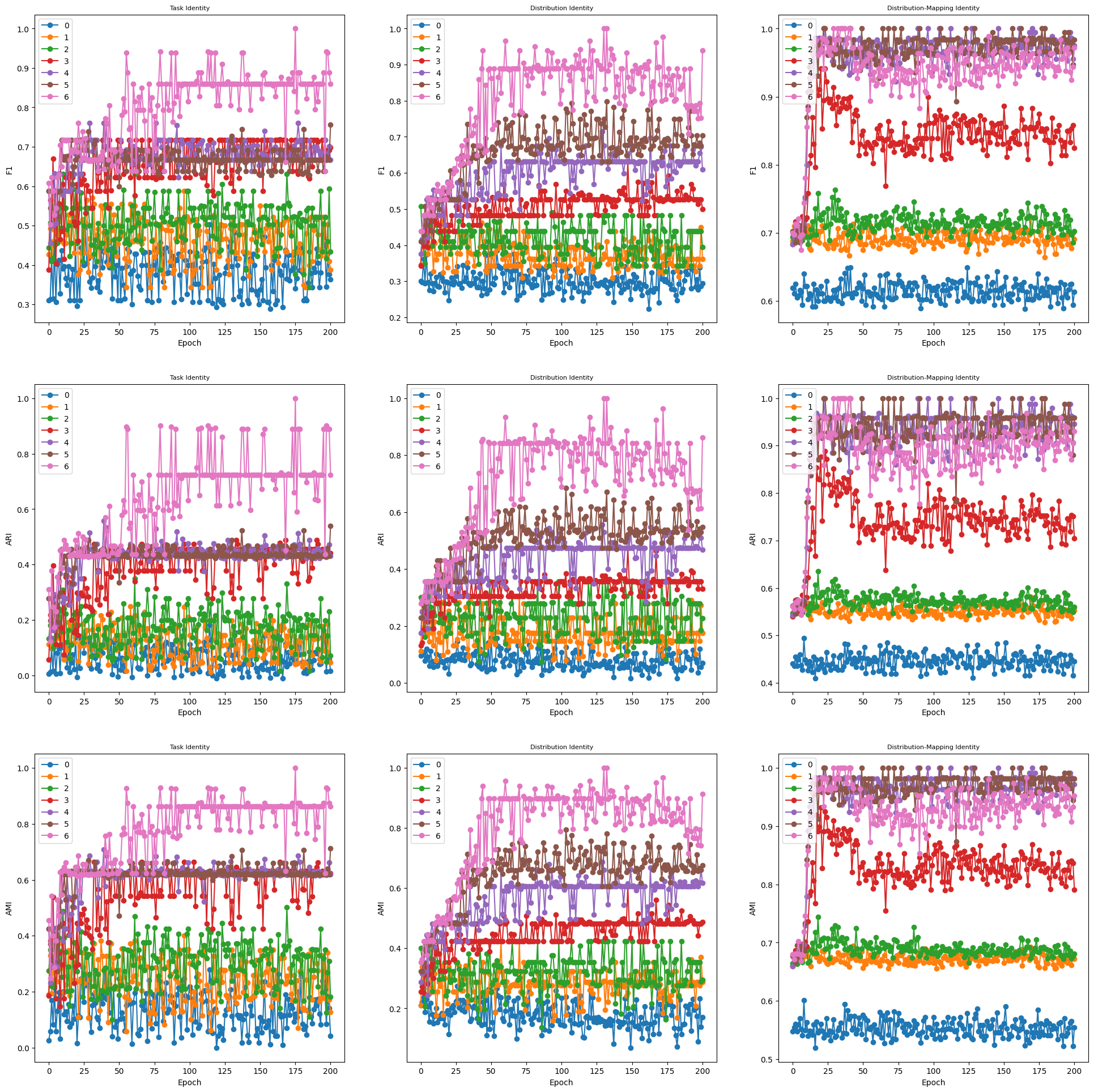}
  \caption{Training subset}
\end{subfigure}%
\\
\begin{subfigure}{\textwidth}\label{fig:syn_all_val}
  \centering
  \includegraphics[width=0.7\linewidth]{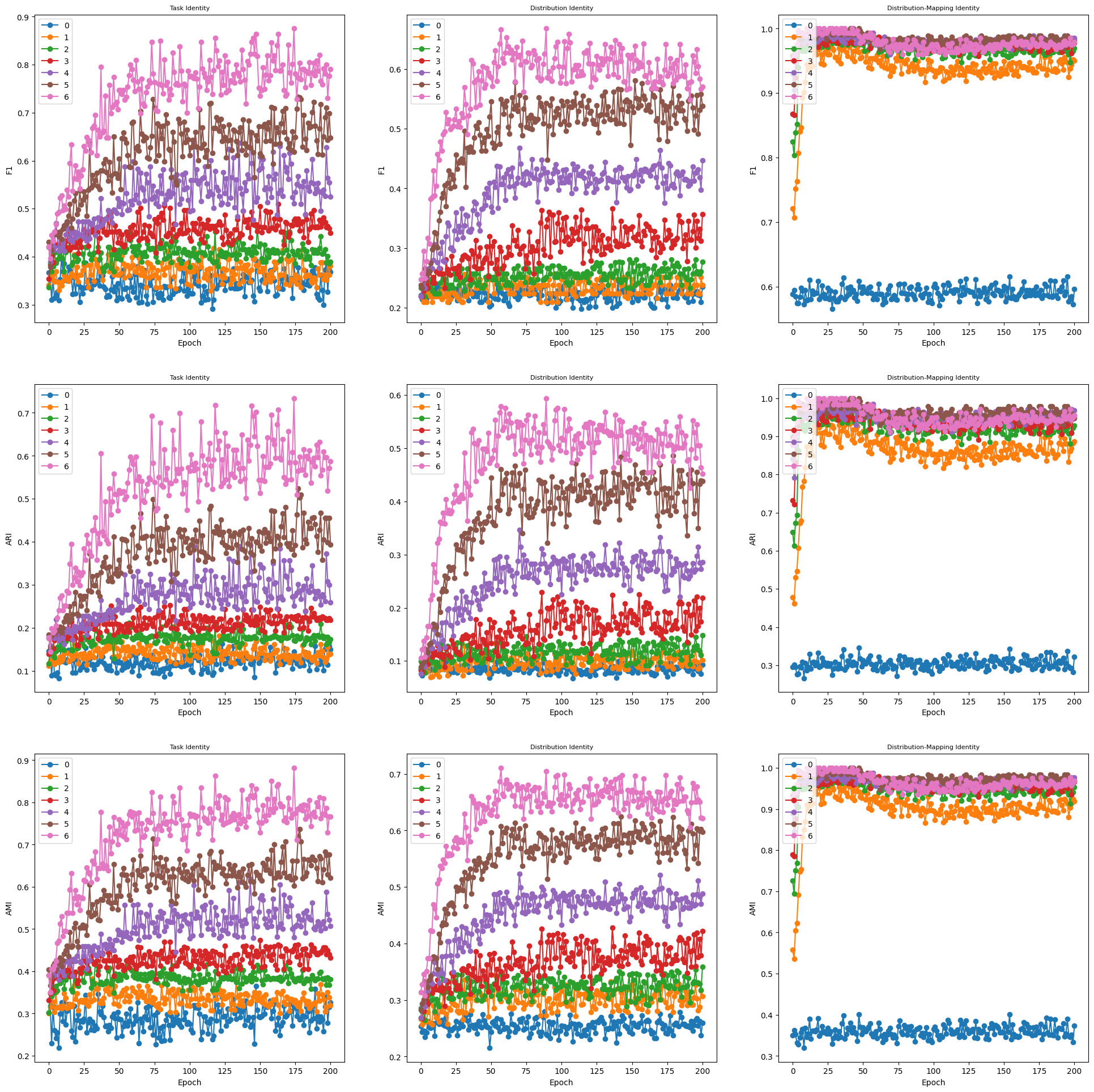}
  \caption{Validation set}
\end{subfigure}%
\caption{Clustering analysis on both training subset (a) and validation set (b) across different layers throughout the training process: Different columns correspond to uses of different identities as labels.}
\label{fig:learning_dynamics_all}
\end{figure*}

\begin{figure*}[!htb]
\centering
\begin{subfigure}{\textwidth}\label{fig:syn_all_trainsub_32}
  \centering
  \includegraphics[width=0.7\linewidth]{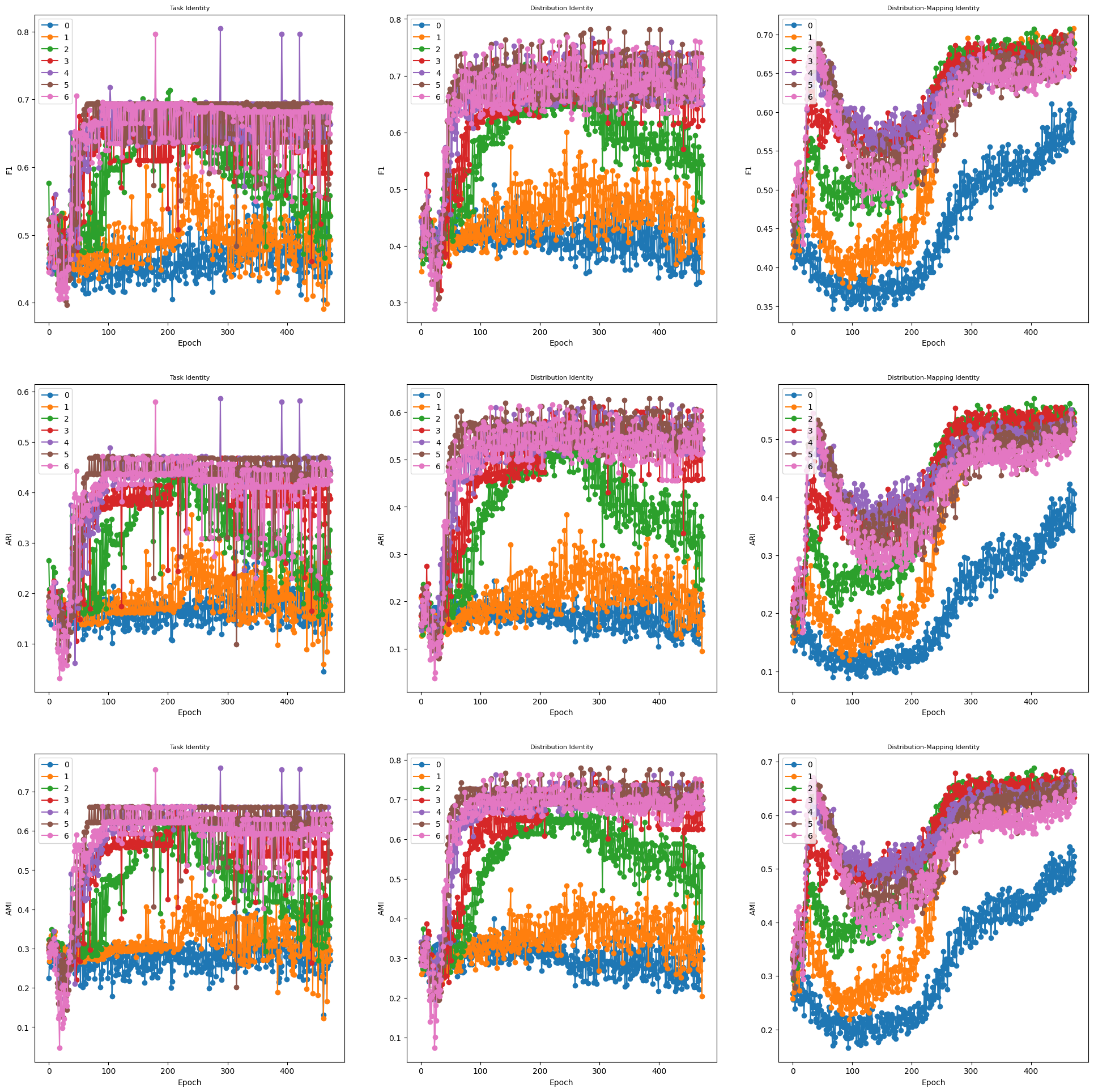}
  \caption{Training subset}
\end{subfigure}%
\\
\begin{subfigure}{\textwidth}\label{fig:syn_all_val_32}
  \centering
  \includegraphics[width=0.7\linewidth]{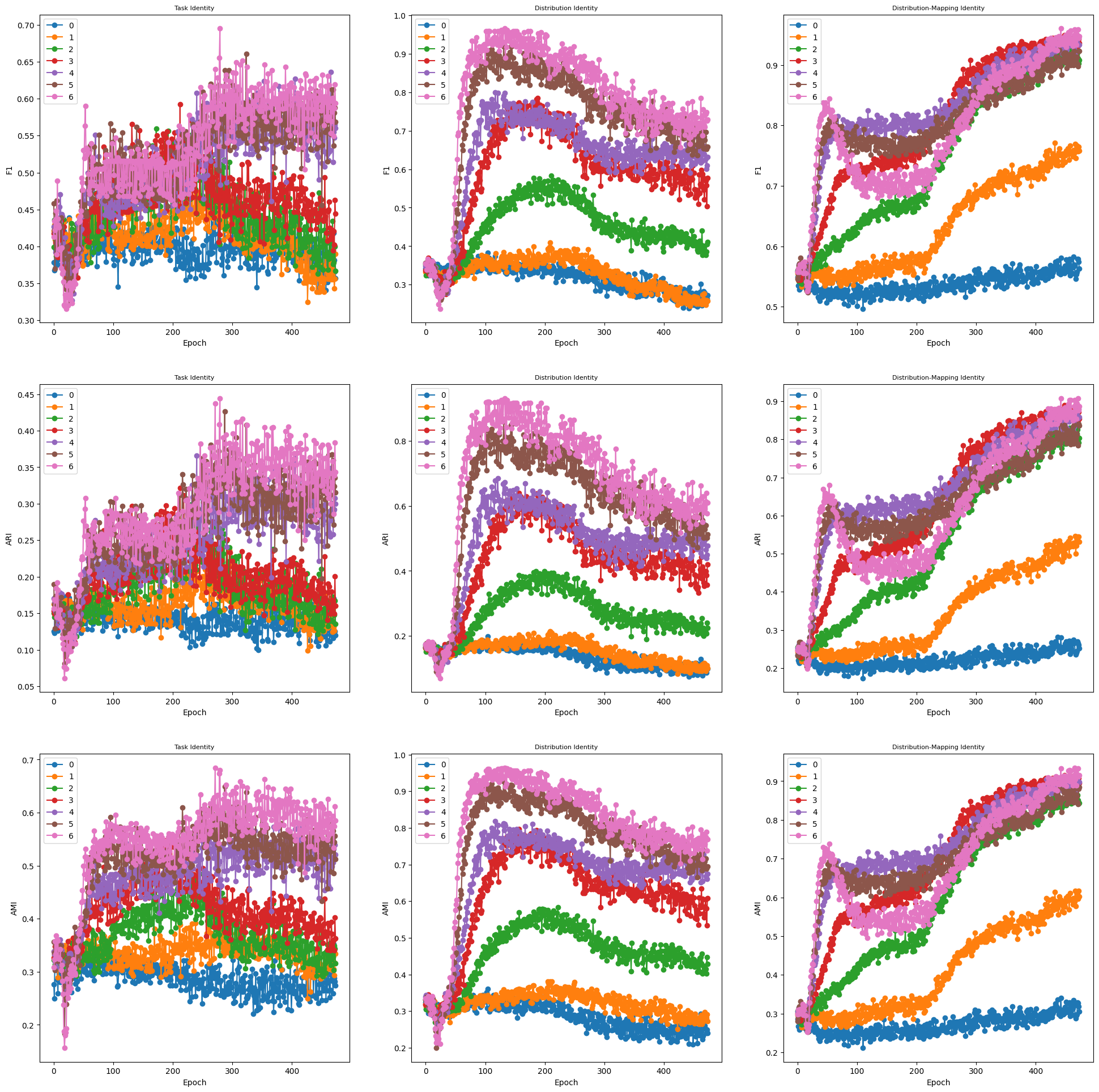}
  \caption{Validation set}
\end{subfigure}%
\caption{Clustering analysis on both training subset (a) and validation set (b) across different layers throughout the training process: The results are shown for the model with 32 hidden dimensions. We train this model for 500 epochs due to its slow convergence. Different columns correspond to the uses of different identities as labels.}
\label{fig:learning_dynamics_all_32}
\end{figure*}

\begin{figure*}[!htb]
\centering
\begin{subfigure}{0.5\textwidth}
  \centering
  \includegraphics[width=0.8\linewidth]{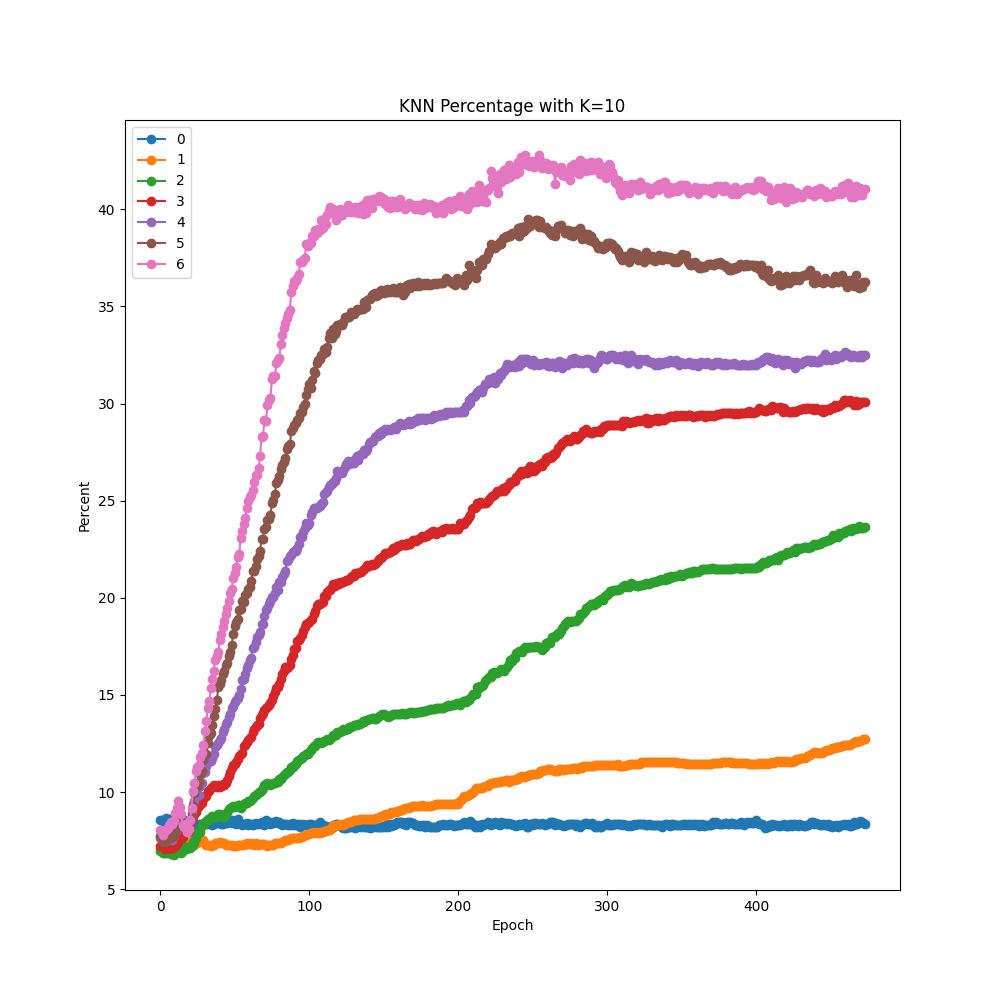}
  \caption{KNN Percentage}
  \label{fig:knn_task_10_32}
\end{subfigure}%
\begin{subfigure}{0.5\textwidth}
  \centering
  \includegraphics[width=0.8\linewidth]{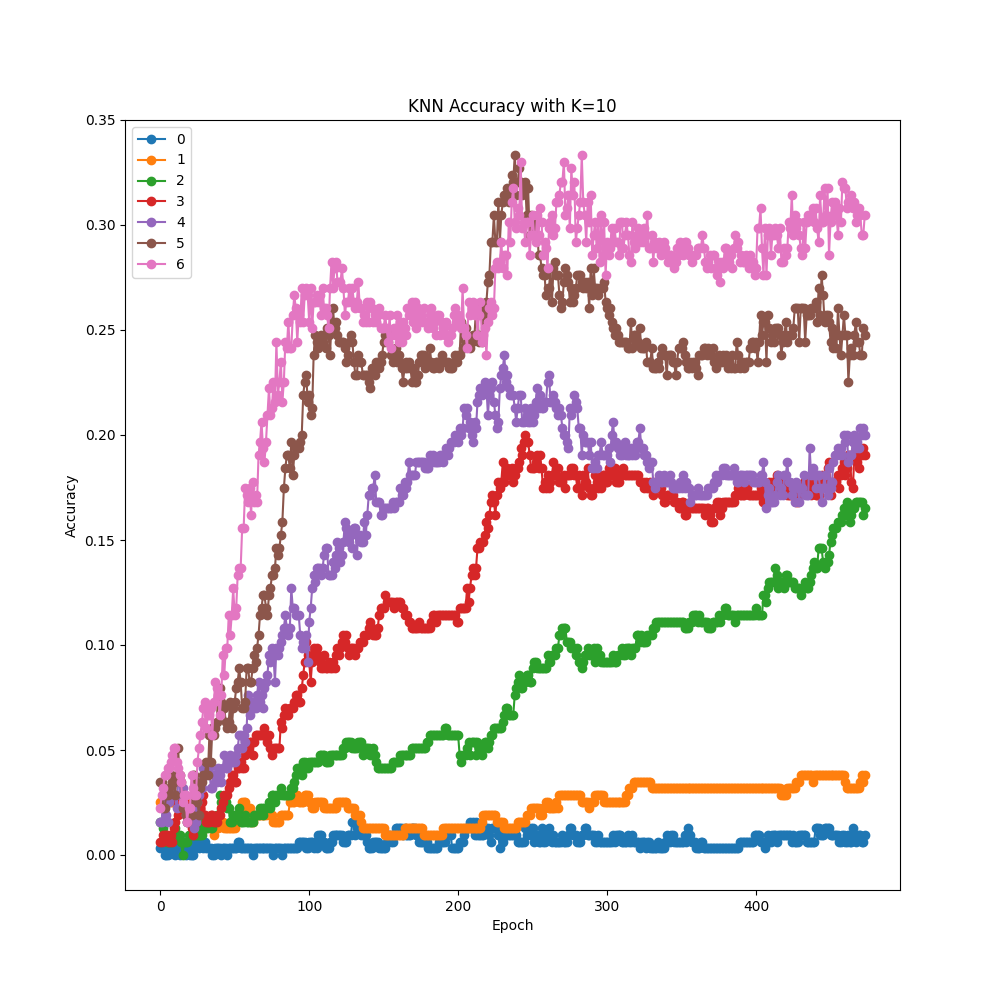}
  \caption{KNN Accuracy}
  \label{fig:knn_acc_10_32}
\end{subfigure}%
\caption{(a) Percentage of K nearest neighbors in the training set of an unseen instance belonging to the same task identity. (b) K nearest neighbors accuracy. Measurements are performed across all of layers and throughout the training process. The results are shown for the model with 32 hidden dimension.}
\label{fig:syn_KNN_32}
\end{figure*}

\begin{figure*}[!htb]
\centering
\begin{subfigure}{\textwidth}\label{fig:syn_all_trainsub_2048}
  \centering
  \includegraphics[width=0.7\linewidth]{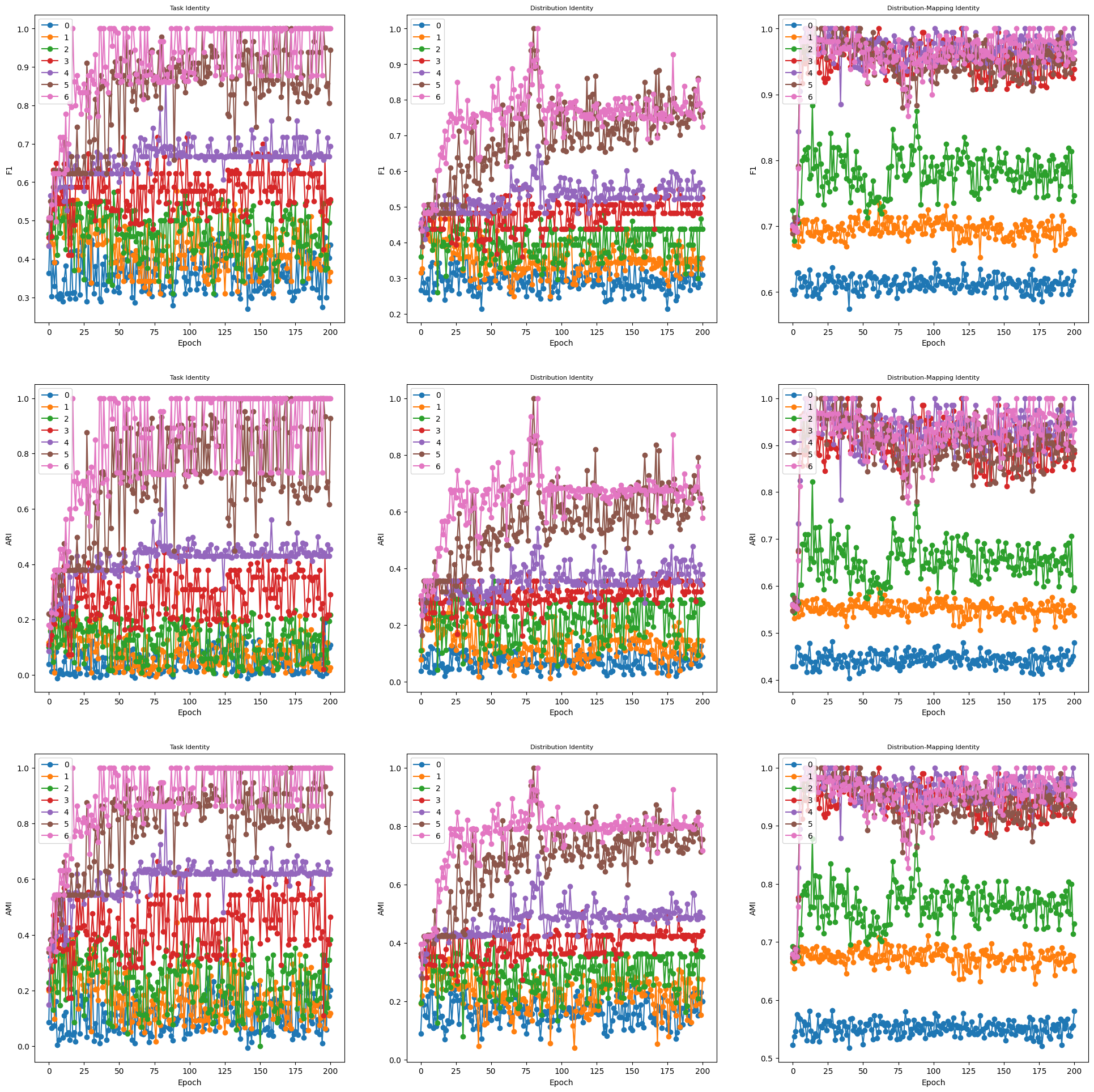}
  \caption{Training subset}
\end{subfigure}%
\\
\begin{subfigure}{\textwidth}\label{fig:syn_all_val_2048}
  \centering
  \includegraphics[width=0.7\linewidth]{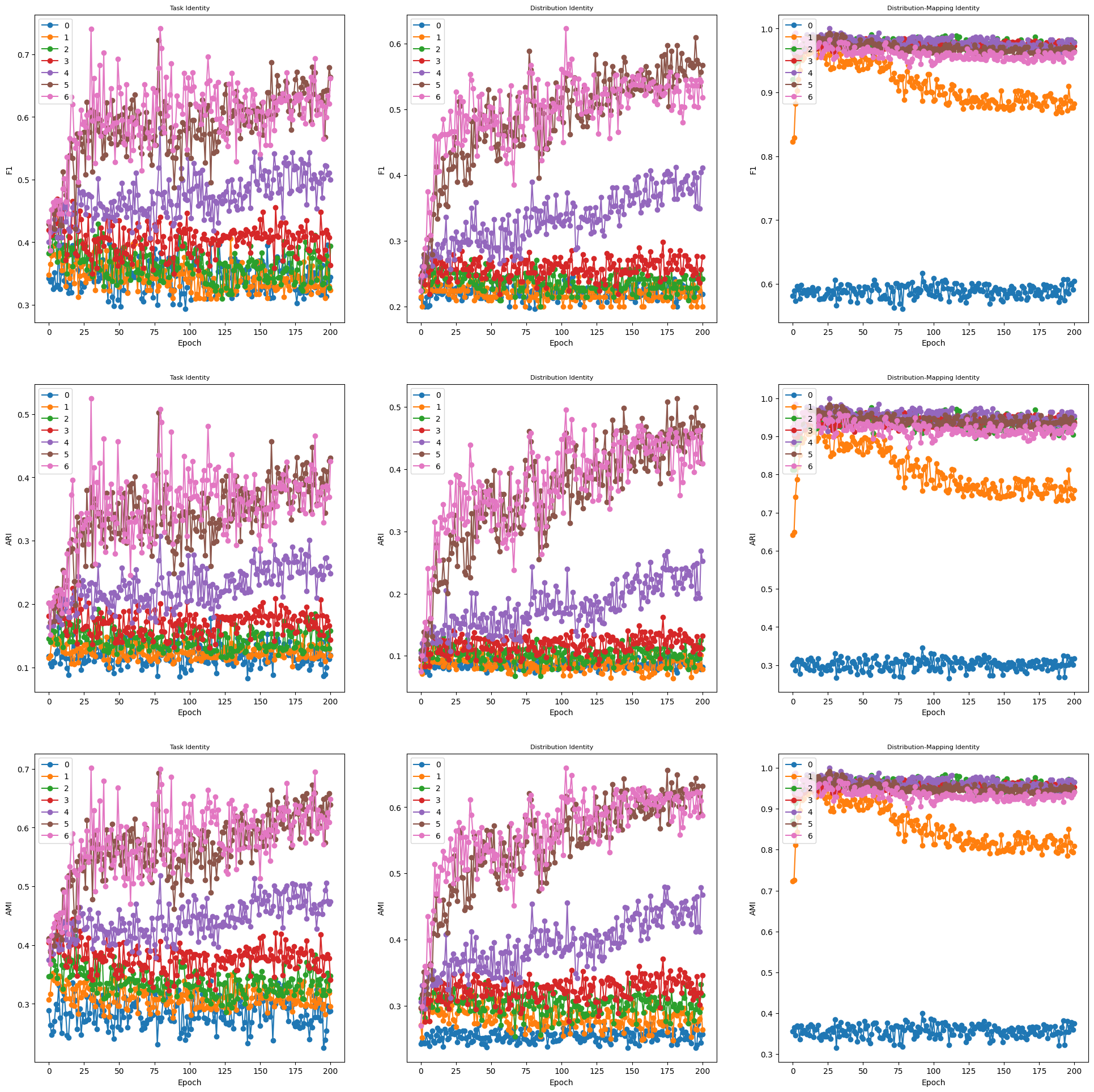}
  \caption{Validation set}
\end{subfigure}%
\caption{Clustering analysis on both training subset (a) and validation set (b) across different layers throughout the training process: The results are shown for the model with 2048 hidden dimension. Different columns correspond to the uses of different identities as labels.}
\label{fig:learning_dynamics_all_2048}
\end{figure*}

\begin{figure*}[!htb]
\centering
\begin{subfigure}{0.5\textwidth}
  \centering
  \includegraphics[width=0.8\linewidth]{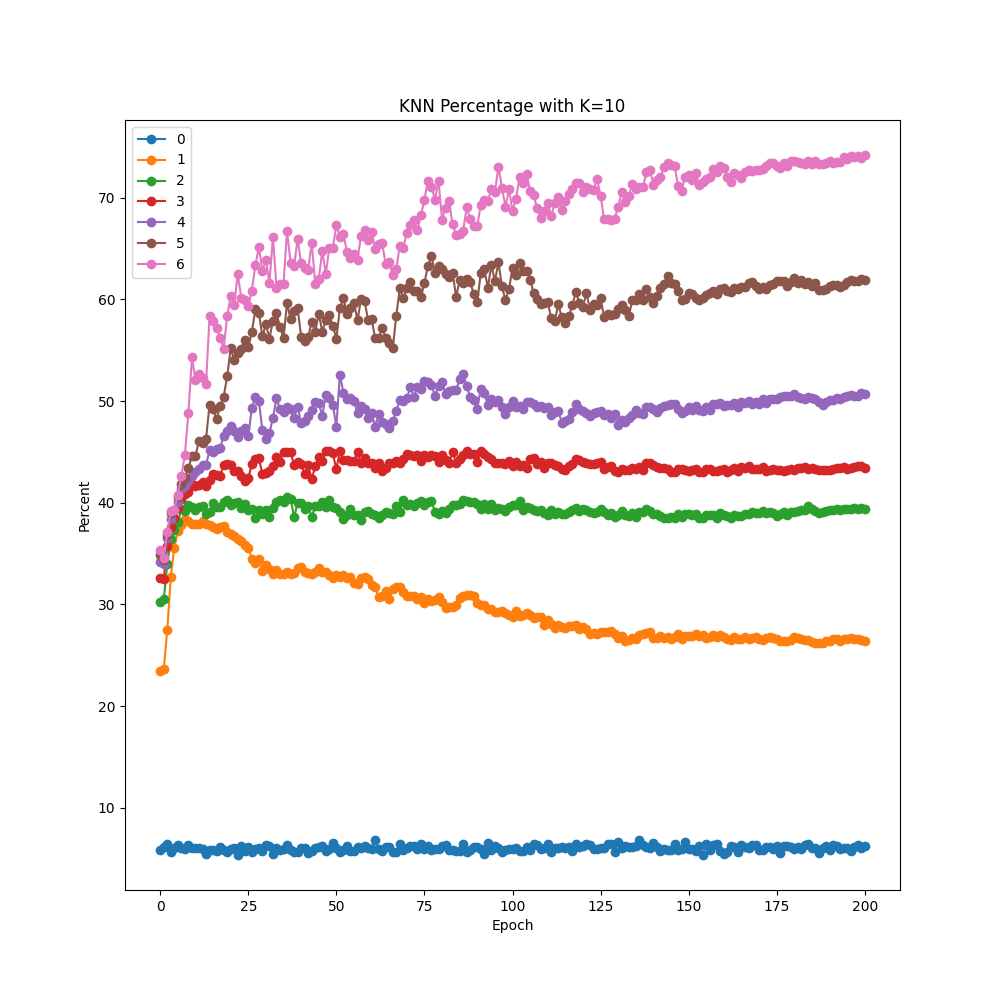}
  \caption{KNN Percentage}
  \label{fig:knn_task_10_2048}
\end{subfigure}%
\begin{subfigure}{0.5\textwidth}
  \centering
  \includegraphics[width=0.8\linewidth]{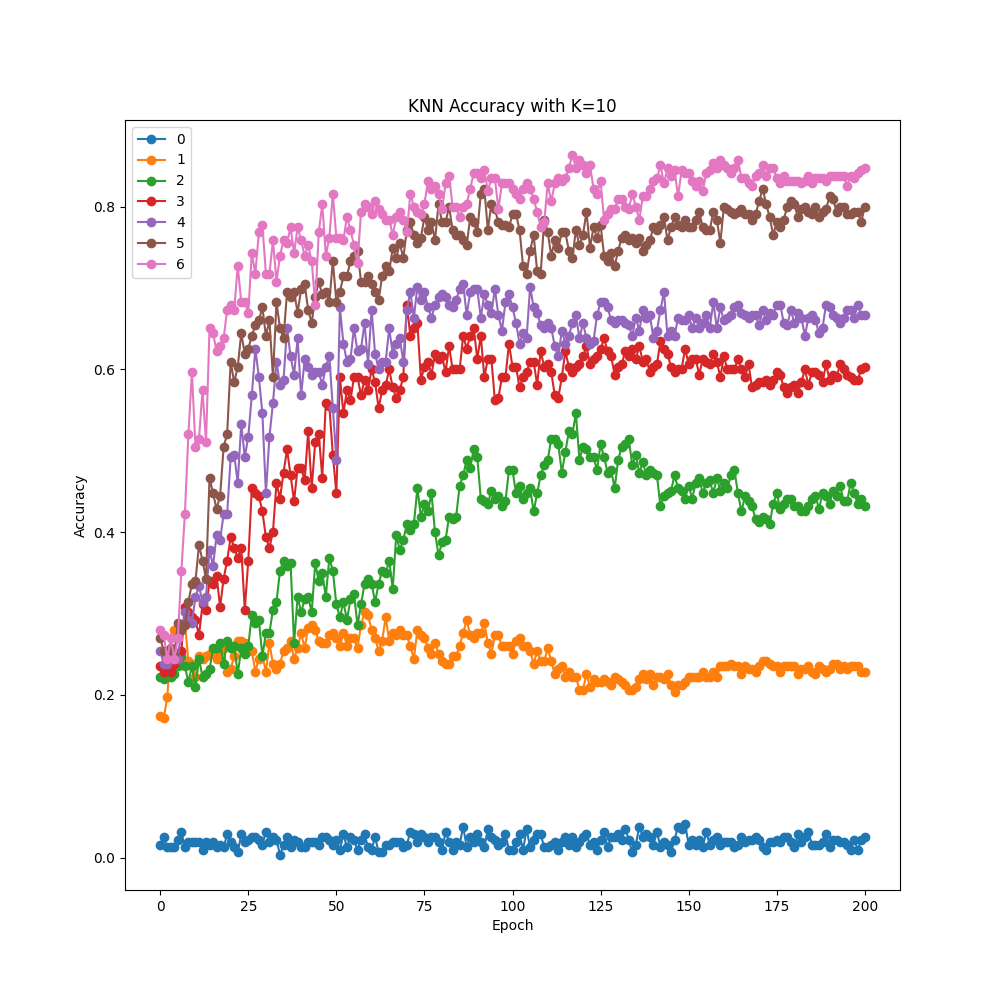}
  \caption{KNN Accuracy}
  \label{fig:knn_acc_10_2048}
\end{subfigure}%
\caption{(a) Percentage of K nearest neighbors in the training set of an unseen instance belonging to the same task identity. (b) K nearest neighbors accuracy. Measurements are performed across all layers and throughout the training process. The results are shown for the model with 2048 hidden dimensions.}
\label{fig:syn_KNN_2048}
\end{figure*}

\end{document}